\newcommand{\cmmnt}[1]{\ignorespaces}
\documentclass[conference,a4paper]{IEEEtran}
\usepackage{graphicx}

\usepackage{multirow}     
\usepackage{booktabs}     
\usepackage[numbers]{natbib}
\usepackage{algorithm}
\usepackage{algpseudocode}
\usepackage{amsmath}      
\usepackage[english]{babel} 
\usepackage{amsthm}         
\usepackage{amsfonts}
\usepackage[flushleft]{threeparttable}
\usepackage{array}
\usepackage{subcaption}
\usepackage{pifont}
\usepackage{mathtools}
\usepackage{graphicx}
\newcommand\scalemath[2]{\scalebox{#1}{\mbox{\ensuremath{\displaystyle #2}}}}

\begin{document}
\title{Assessing the Impacts of Imperfect Datasets on Client Selections in Federated Learning}
\author{\IEEEauthorblockN{Yuan-Heng Tsai, Li-Hsing Yen, and Yan-Wei Chen}
\IEEEauthorblockA{\textit{Department of Computer Science, National Yang Ming Chiao Tung University, Hsinchu, Taiwan.}
}}

\maketitle

\begin{abstract}
Federated learning (FL) is a popular distributed learning framework where multiple clients perform local training and a server aggregates the locally updated models. FL enables decentralized training while preserving the privacy of clients' datasets. However, non-independent and identically distributed (non-IID) or noisy datasets can lead to low model accuracy or high convergence latency. Precluding these clients through client selection may mitigate the problem, but heavily biased client selections may also degrade the learning performance. In this study, we first experimentally measure the impact of non-IID data (including skews in data quantity and label distribution), noisy data, and fairness in client selection on model accuracy and convergence. We then propose a privacy-preserving scoring method to assess each client's contribution in FL, with experiments conducted to demonstrate the effectiveness of the proposed assessment. 

\end{abstract}
\section{Introduction}
Federated learning (FL) enables decentralized model training by FL clients while ensuring the clients' privacy. 
An FL training task begins with a central server initializing a global model and sending it to all clients. Each client then uses its local data to train the model, computes an update such as gradients or model parameters, and sends back the local update to the server. The server aggregates local updates to update the global model and then redistributes the updated global model to all clients. This completes one round of FL and the process repeats until convergence.

FL is challenged by several issues. Particularly, the global model may not perform well when clients have varying data distributions known as non-independent and identically distributed (\emph{non-IID}) data. 
Non-IID data refers to significant skews in label, feature, or quantity distributions in the dataset. It can deteriorate the model accuracy and increase convergence latency \cite{HH+22}, particularly when clients have severe non-IID datasets. Another challenge comes from noisy data or mislabeled samples, which can also reduce the quality of the aggregated model. 

Some studies mitigate statistical heterogeneity by precluding or down-weighting clients with highly non-IID data to stabilize convergence \cite{ZLL+18,CWJ22}.
In contrast, recent fairness-aware approaches argue that promoting balanced or fair client participation can also enhance convergence and improve overall model accuracy \cite{SLS+23}.
Moreover, it is still an open question how to predict a client's contribution to FL under this circumstance without privacy leakage. Addressing this question can help design a sophisticated client selection scheme.   

As most prior research focuses on partial aspects of this issue, we comprehensively analyze the challenges of non-IID and noisy datasets. We also study experimentally whether fairness in client selections can significantly affect the model quality and convergence time. We then present a scoring method to assess each client's importance in client selections considering both dataset quality and fairness. 

The rest of this paper is organized as follows. Sec.~\ref{se:relatedwork} introduces the background and related work on FL. We present experiments in Sec. 3 to study the impact of non-IID datasets and unfair client selections on test accuracy and convergence. We propose an assessment method to measure each client’s contribution without breaching privacy in Sec. 4. Sec. 5 concludes this paper.

\section{Background and Related Work}
\label{se:relatedwork}


\subsection{Non-IID Datasets in Federated Learning}
Many studies reported that non-IID data degrades the model performance and convergence time. The non-IID setting can fall into several classes: quantity skew, label skew, and  feature skew \cite{ZXL+21,ZLL+18}. 

\subsubsection{Quantity Skew} 
\emph{Quantity skew} refers to a skewed distribution in sample quantity across clients.
In FL, if data are IID across clients and the server uses FedAvg ~\cite{MMR+17} to weight each client’s update by its dataset size, then the expected update of FedAvg equals the centralized gradient update on the union of all data \cite{MMR+17}. 
So the global model's convergence and final accuracy depend only on the total dataset size, not on how those samples are distributed across clients.
Therefore, quantity skew does not pose a serious problem for client selection as long as the server is aware of the distribution of sample quantities across clients.
With that information, many client selection methods tend to favor clients with larger datasets \cite{CWJ22}.

\subsubsection{Label Skew}
\emph{Label skew} refers to a skewed (imbalanced) distribution in label across clients, 
which can cause significant model weight divergence \cite{ZLL+18}. 
A study \cite{DLH+23} recommends augmenting local datasets with extra samples to balance the label distributions. 
Other approaches have been proposed to prevent aggregating highly divergent models due to label skew.
Examples are Personalized Federated Learning (PFL) \cite{MHM+20,HCZ+20} and Clustered Federated Learning \cite{IJX+24,LXS+22}.
Some studies \cite{NNN+22,ZLT+22} reduced the impact of selecting clients with highly label-skewed datasets by lowering their weights in the aggregation. 
Many approaches have employed deep reinforcement learning to optimize client selection in the presence of label-skewed or otherwise non-IID data distributions \cite{NNN+22,MLL+23}.

\subsubsection{Feature Skew}
\emph{Feature skew} refers to the situation where clients possess data with different feature distributions. For instance, in the MNIST dataset \cite{mnist}, a digit might be handwritten in various styles, while in the CIFAR-10 dataset \cite{cifar10}, animals could be photographed from different angles. This results in unique feature sets for the same labels. 
Recent studies \cite{ZXL+21,ZLL+18} show that feature-skew is a significant non-IID issue in FL, characterized by varying relationships between features and labels across different clients.  
Different from label skew, which can be statistically analyzed, feature skew poses a more significant challenge for quantification due to its ambiguous definitions and widespread occurrence.

In light of this, we do not view feature skew negatively but as an essential element of data diversity that ensures effective model generalization across diverse feature representations within the same label. Therefore, it is crucial to implement fairness in client selection. Such fairness ensures that the model adapts to and benefits from the intrinsic variations within the data, enhancing its robustness and general applicability.

\subsection{Mislabeled Samples}
Client datasets may contain mislabeled samples that significantly degrade the performance of the global model. 
For deliberate poison attacks, some studies \cite{FYB+20} used the similarity between the local update gradient and the global update gradient to gauge the extent of mislabeling. 
For noisy labels, some studies \cite{YPB+20,LLC+23} detected and filtered out mislabeled samples from local datasets. 
Another way to mitigate mislabeling is to add additional information to the local datasets \cite{XCQ+22,ZYC+22}. 
If clients can be trusted, the server can measure clients' mislabel levels based on a pre-trained model to evaluate the local models \cite{YQW+21,DLR+21,TSO+24}.
Many studies \cite{FYB+20,YQW+21,TSO+24}  
mitigated the impact of mislabeling by adjusting the aggregation weights for clients that may contain mislabels. 



\subsection{Fairness in Client Selections}
An uneven selection can cause the global model to learn predominantly from a small subset of clients, resulting in rapid convergence to an undesired local optimum.  
Therefore, \emph{fairness}, which ensures that all clients have enough opportunities to participate, can potentially enhance model performance \cite{SYL+21}. 
This is why some studies considered both effective participation and fairness in client selections \cite{HLS+21}. 
Many studies \cite{HLW+21,BLM+22,ZZQ+22} implemented long-term fairness constraints and used Lyapunov optimization to ensure each client's average participation rate.

\section{Impact of Imperfect Data and Unfair Client Selections}
\label{se:pre-experiment} 
We conducted several experiments to study how imperfect data (including skewed distributions in label and sample quantity and datasets containing mislabeled samples) as well as unfair client selections affect the model performance and convergence speed.

\begin{table}[tb]
\centering
\caption{Parameter Setting}
\label{table:setting}
\begin{tabular}{ccl}
\toprule
Parameter & Value & Description\\
\midrule
$|C|$ & 100 & Total number of clients\\
$N_\text{max}$ & $10$ & Maximum number of clients to be selected\\
$n$ & 10 & Number of labels\\
$I_\text{loc}$ & $10$ & Number of local epochs\\
$B_\text{loc}$ & $64$ & Training batch size\\
$\gamma_\text{loc}$ & $0.005$ & Local training learning rate\\
\bottomrule					 
\end{tabular}
\end{table}

We tested two classic datasets: MNIST and CIFAR-10.
Unless otherwise specified, there were 100 clients, each with an IID dataset with 200 samples in the following experiment settings. 
We used the FedAvg algorithm with parameter setting shown in Table~\ref{table:setting}. 
In each round, 10 out of 100 clients were randomly selected for participation in the training.
Each result was averaged over ten experiments.

\subsection{Impact of Dataset Sizes}
We tested five different dataset sizes (number of samples in a dataset). All clients possessed datasets of equal size. 
The results 
confirm that larger dataset sizes correlated with improved test accuracy and faster convergence, which is consistent with the observations of previous studies and centralized learning mechanisms. Additionally, datasets with sizes below a certain threshold failed to support effective learning. 

\subsection{Impact of Label Skews}

\begin{table}[tb]
\centering
\caption{Six Scenarios of Label Skews}
\label{table:scenario_ls}
\begin{tabular}{l p{2.4in}}
\toprule
Name & Description\\
\midrule
IID & Each client possessed an equal distribution of all labels. (no label skew)\\
SingleLabel & Each client exclusively possessed samples from a single label. (extreme label skew)\\
TwoLabels & Each client had samples from two labels with an equal sample count for both labels.\\
70\%TwoLabels & 30\% of clients followed the IID setting and the remaining 70\% followed the TwoLabels setting.\\
50\%TwoLabels & 50\% of clients used IID settings and 50\% used the TwoLabels settings.\\
30\%TwoLabels & 70\% of clients used IID settings and 30\% used the TwoLabels setting.\\
\bottomrule					 
\end{tabular}
\end{table}

We explored six scenarios with varying degrees of label skews (Table~\ref{table:scenario_ls}) to see how label skew affects model accuracy.
\begin{figure}[tb]
    \centering
    \begin{subfigure}{0.48\linewidth}
        \centering
        \includegraphics[width=\textwidth]{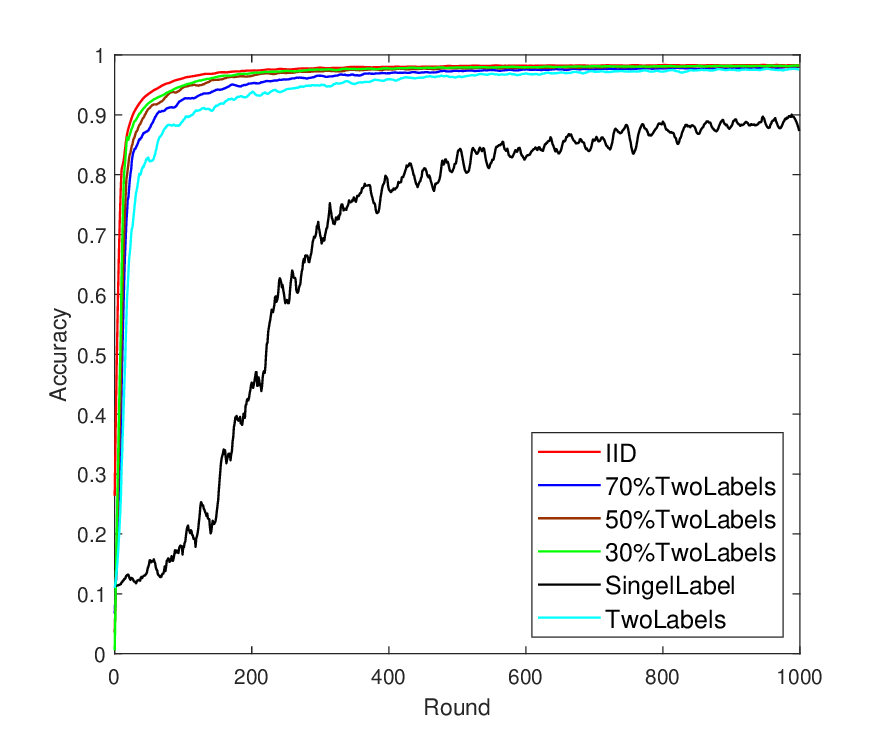}
        \caption{MNIST}
        \label{fig:pre-labelskew-impact-mnist}
        
    \end{subfigure}%
    \begin{subfigure}{0.48\linewidth}
        \centering
        \includegraphics[width=\textwidth]{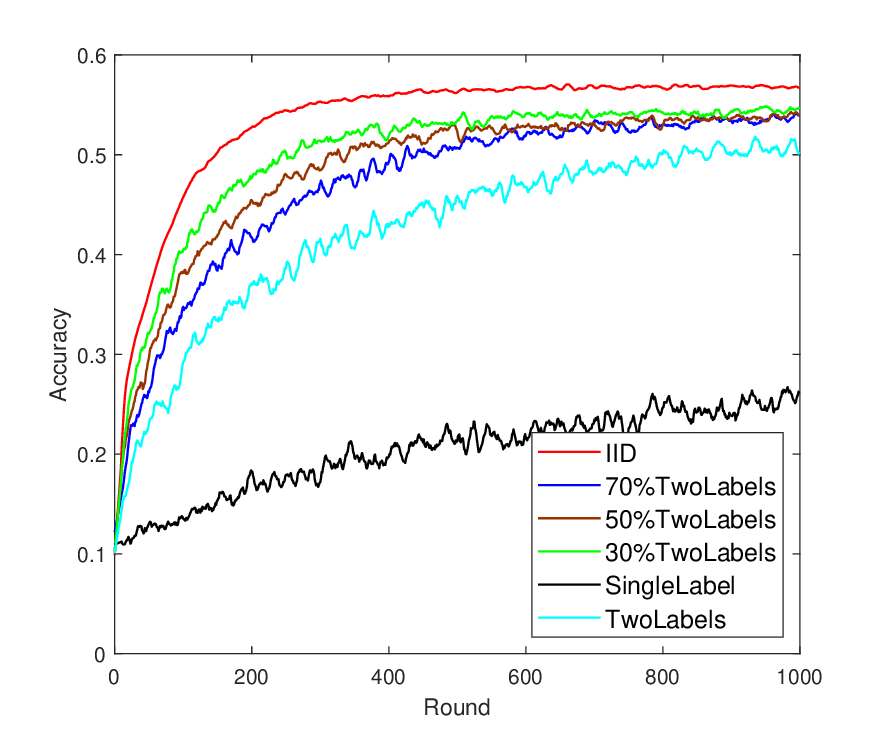}
        \caption{CIFAR-10}
        \label{fig:pre-labelskew-impact-cifar10}
    \end{subfigure}
    \caption{Test accuracy over training rounds under different settings of label skew}
    \label{fig:pre-labelskew-impact}
\end{figure}
The result in Fig.~\ref{fig:pre-labelskew-impact-cifar10} from the CIFAR-10 dataset revealed a clear trend that, as the degree of label skews increased, both test accuracy and convergence speed worsened. By contrast, the result in Fig.~\ref{fig:pre-labelskew-impact-mnist} from the MNIST dataset shows only slight performance declines under the same conditions, except in the extreme case of SingleLabel. This revealed that different tasks had varying levels of resilience to label skew.

\subsection{Impact of Mislabeled Samples}
\label{impact:mislabel}
\begin{figure}[tb]
    \centering
    \begin{subfigure}{0.48\linewidth}
        \centering
        \includegraphics[width=\textwidth]{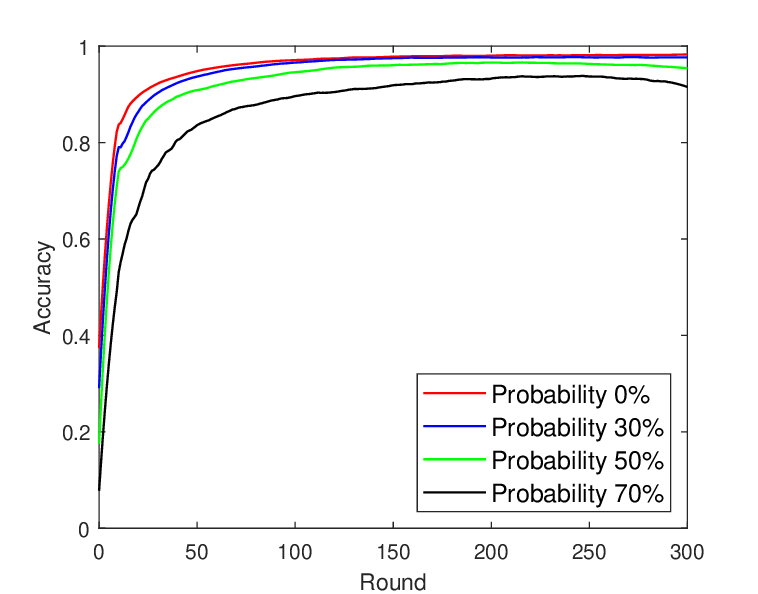}
        \caption{MNIST}
        \label{fig:pre-mnist-mislabel-random-impact}
    \end{subfigure}
    \begin{subfigure}{0.48\linewidth}
        \centering
        \includegraphics[width=\textwidth]{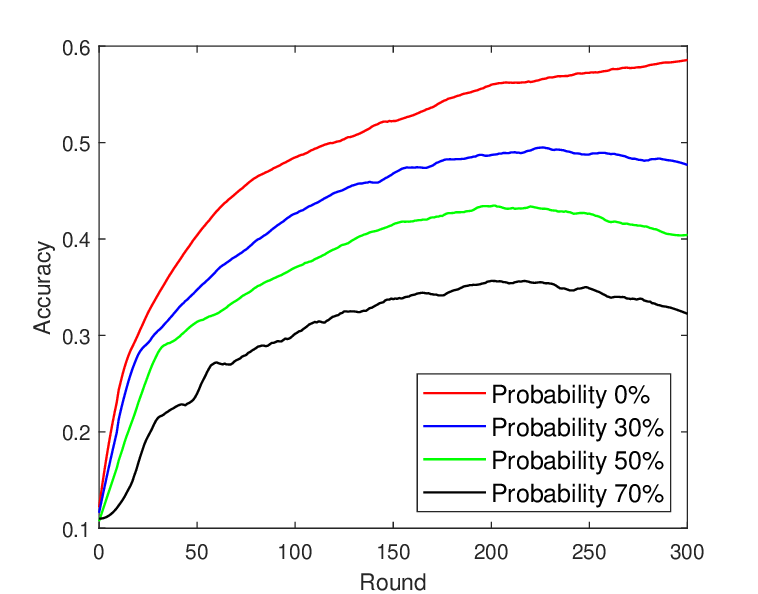}
        \caption{CIFAR-10}
        \label{fig:pre-cifar10-mislabel-random-impact}
    \end{subfigure}
    \caption{Test accuracy over training rounds under different probabilities of random mislabeling}
    \label{fig:pre-mislabel-random-impact}
\end{figure}

\begin{figure}[tb]
    \centering
    \begin{subfigure}{0.48\linewidth}
        \includegraphics[width=\linewidth]{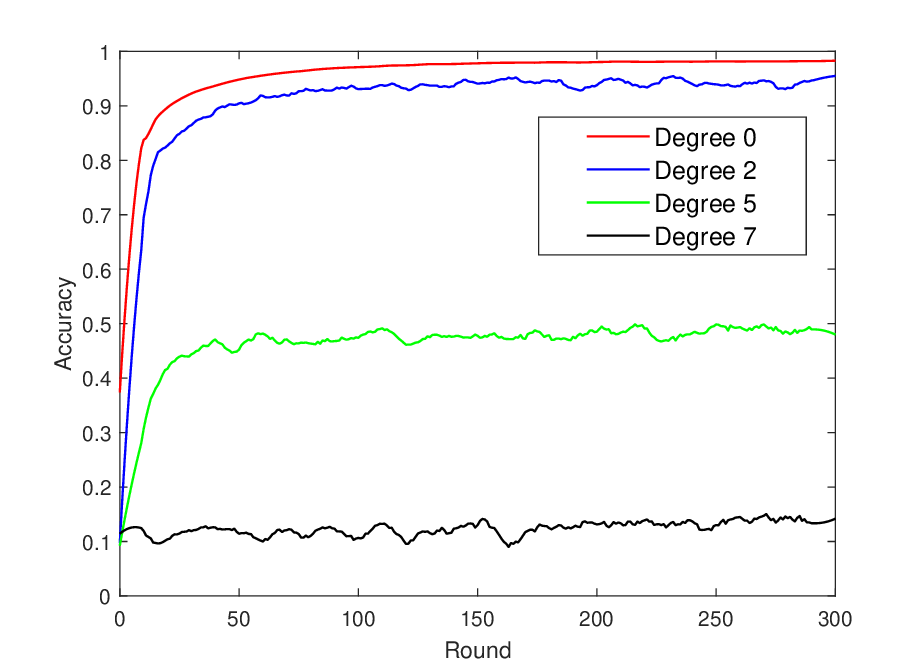}
        \caption{MNIST}
        \label{fig:pre-mnist-mislabel-sequential-impact}
    \end{subfigure}%
    \begin{subfigure}{0.48\linewidth}
        \includegraphics[width=\textwidth]{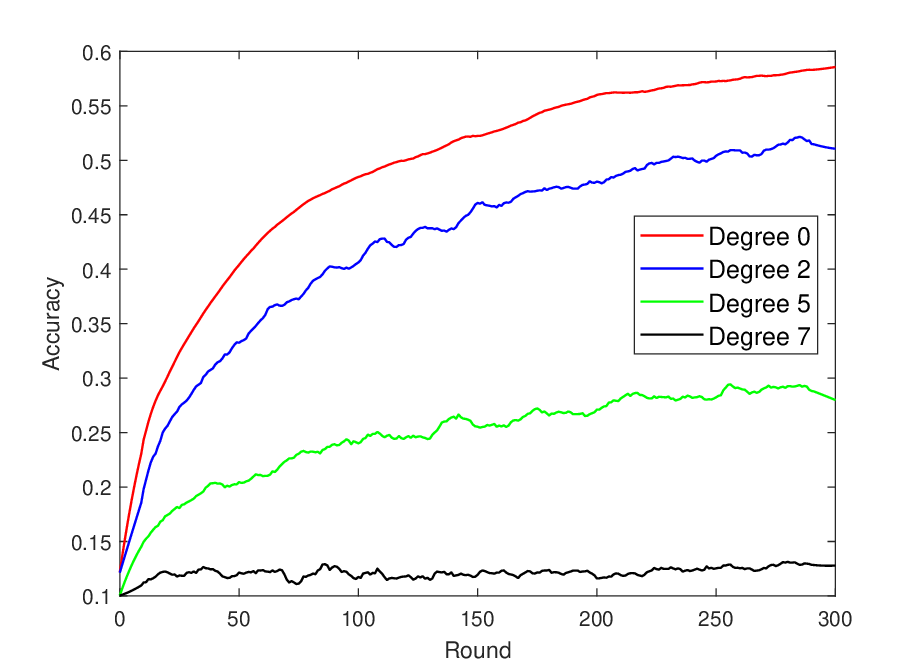}
        \caption{CIFAR-10}
        \label{fig:pre-cifar10-mislabel-sequential-impact}
    \end{subfigure}
    \caption{Test accuracy over training rounds under different degrees of sequential mislabeling}
    \label{fig:pre-mislabel-sequential-impact}
\end{figure}

\begin{figure}[tb]
    \centering
    \begin{subfigure}{0.48\linewidth}
        \centering
        \includegraphics[width=\textwidth]{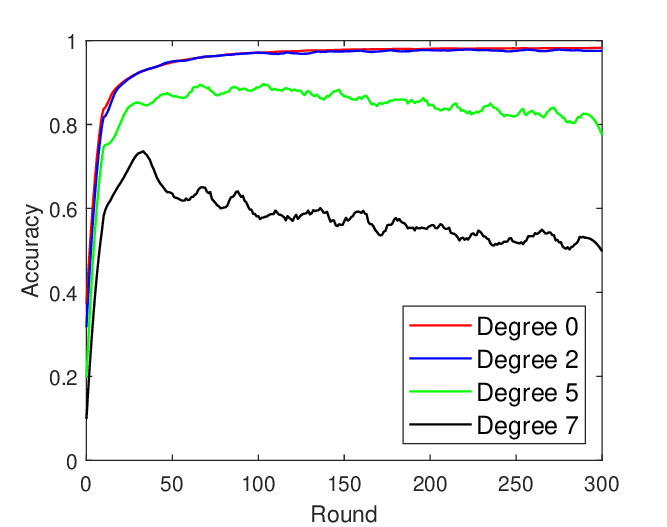}
        \caption{MNIST}
        \label{fig:pre-mnist-mislabel-cyclic-impact}
    \end{subfigure}%
    \begin{subfigure}{0.48\linewidth}
        \centering
        \includegraphics[width=\textwidth]{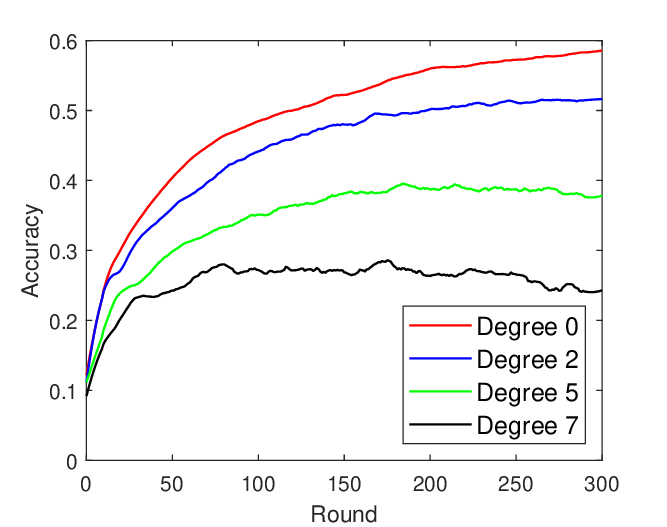}
        \caption{CIFAR-10}
        \label{fig:pre-cifar10-mislabel-cyclic-impact}
    \end{subfigure}
    \caption{Test accuracy over training rounds under different degrees of cyclic mislabeling (with $90$ clients)}
    \label{fig:pre-mislabel-cyclic-impact}
\end{figure}

We explored three common mislabeling scenarios mentioned in the literature.
The first is \emph{random mislabeling} with probability $y$, where every sample had a probability of $y$ being mislabeled with a random label. Fig.~\ref{fig:pre-mislabel-random-impact} shows the test accuracy over training rounds under different values of $y$.
The second scenario is \emph{sequential mislabeling} with degree $y$, where some labels were relabeled to the subsequent labels and the degree $y$ specifies the number of such labels. For example, when $y=k$, all samples with labels $1, 2, ..., k$ were relabeled as labels $2, 3, ... k+1$, respectively. 
Fig.~\ref{fig:pre-mislabel-sequential-impact} shows the result with some values of $y$.
The third scenario is \emph{cyclic mislabeling} with degree $y$, which  
is similar to sequential mislabeling except that all samples with label $k$ were relabeled as label 1 when $y=k$. 
Fig.~\ref{fig:pre-mislabel-cyclic-impact} shows the result with some values of $y$.

From these results, we note a marked decline in performance correlating with increased probability or degree of mislabeling. Compared with the results associated with label skews (Fig.~\ref{fig:pre-labelskew-impact}), where learning was attainable albeit at a reduced efficiency, the results here demonstrated a pronounced drop in performance. This was especially evident in sequential mislabeling with $y=7$ (Fig.~\ref{fig:pre-mislabel-sequential-impact}), where the learning process failed. This observation suggests that client selections should exclude clients suffering from severe mislabeling.

Among the three mislabeling scenarios, the sequential mislabeling scenario shown in Fig.~\ref{fig:pre-mislabel-sequential-impact} consistently resulted in the most significant performance degradation, highlighting its potential as a stress test for further refining our approach. Therefore, we will focus on the sequential mislabeling scenario in our upcoming experiments.

\subsection{Impact of Unfair Client Selections} \label{pre_exp_fairness}
We explored how fairness in client selection affects the model's performance and convergence speed. To understand this, we conducted experiments on datasets with skewed label distributions and mislabeled samples.

\begin{figure}[tb]
    \centering
    \begin{subfigure}{0.32\linewidth}
        \centering
        \includegraphics[width=\textwidth]{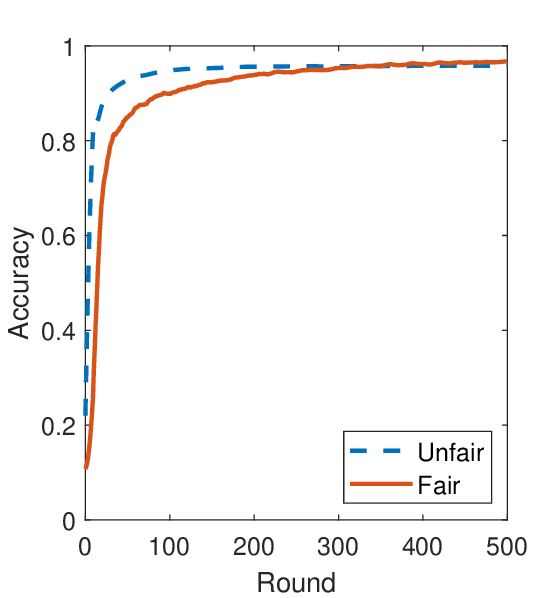}
        \caption{Two-Label Non-IID}
        \label{fig:pre-fairness-impact-labelskew-mnist-2}
    \end{subfigure}%
    \begin{subfigure}{0.32\linewidth}
        \centering
        \includegraphics[width=\textwidth]{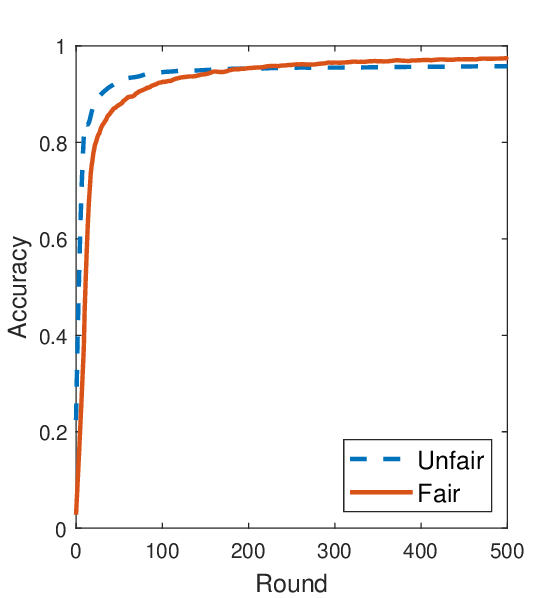}
        \caption{Four-Label Non-IID}
        \label{fig:pre-fairness-impact-labelskew-mnist-4}
    \end{subfigure}
        \begin{subfigure}{0.32\linewidth}
        \centering
        \includegraphics[width=\textwidth]{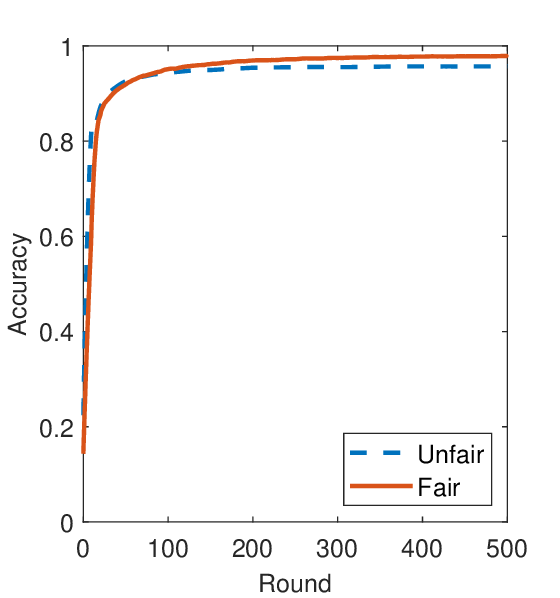}
        \caption{Seven-Label Non-IID}
        \label{fig:pre-fairness-impact-labelskew-mnist-7}
    \end{subfigure}
    \caption{Test accuracy over training rounds under fair vs. unfair client selection with label-skewed datasets (MNIST)}
    \label{fig:pre-fairness-impact-labelskew-mnist}
\end{figure}

\begin{figure}[tb]
    \centering
    \begin{subfigure}{0.32\linewidth}
        \centering
        \includegraphics[width=\textwidth]{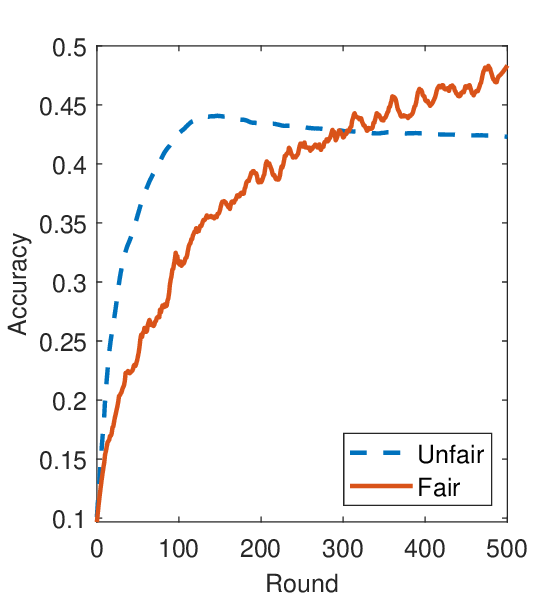}
        \caption{Two-Label Non-IID}
        \label{fig:pre-fairness-impact-labelskew-cifar10-2}
    \end{subfigure}%
    \begin{subfigure}{0.32\linewidth}
        \centering
        \includegraphics[width=\textwidth]{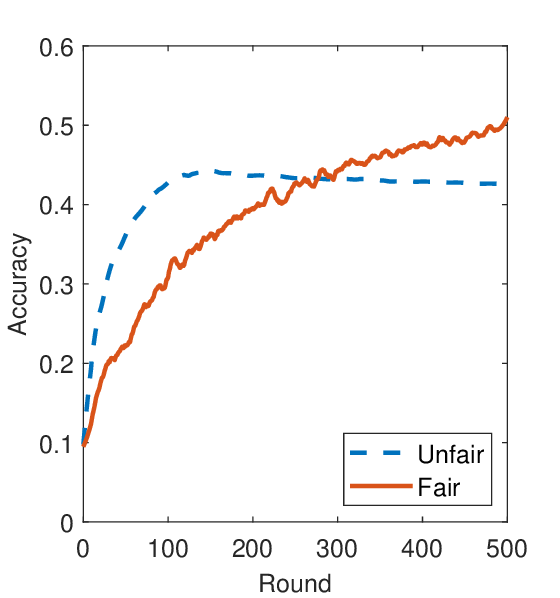}
        \caption{Four-Label Non-IID}
        \label{fig:pre-fairness-impact-labelskew-cifar10-4}
    \end{subfigure}
        \begin{subfigure}{0.32\linewidth}
        \centering
        \includegraphics[width=\textwidth]{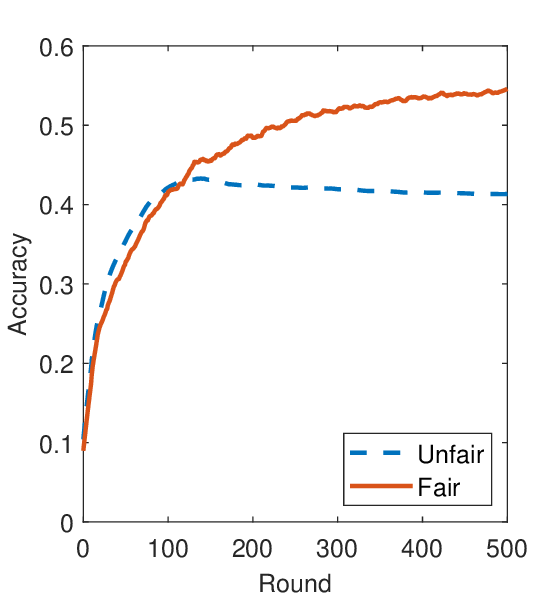}
        \caption{Seven-Label Non-IID}
        \label{fig:pre-fairness-impact-labelskew-cifar10-7}
    \end{subfigure}
    \caption{Test accuracy over training rounds under fair vs. unfair client selection with label-skewed datasets (CIFAR-10)}
    \label{fig:pre-fairness-impact-labelskew-cifar10}
\end{figure}

For skewed label distribution, $90\%$ clients had IID datasets while the rest had samples from only two, four, or seven labels (with an equal sample count for each label). We tested two client selection policies. One policy, labeled Fair in the figures, randomly selected clients, which resulted in nearly uniform selection frequencies across clients. The other policy, Unfair, excluded all the clients with label-skewed datasets in the random client selections. 
Figs.~\ref{fig:pre-fairness-impact-labelskew-mnist} and \ref{fig:pre-fairness-impact-labelskew-cifar10} show the results for MNIST and CIFAR10, respectively.

We observe that fair client selections significantly improved accuracy. However, more global rounds were required to achieve convergence, and the importance of fairness increased as more labels were contained in non-IID datasets. 

\begin{figure}[tb]
    \centering
	\small
    \begin{subfigure}{0.24\linewidth}
        \centering
        \includegraphics[width=\textwidth]{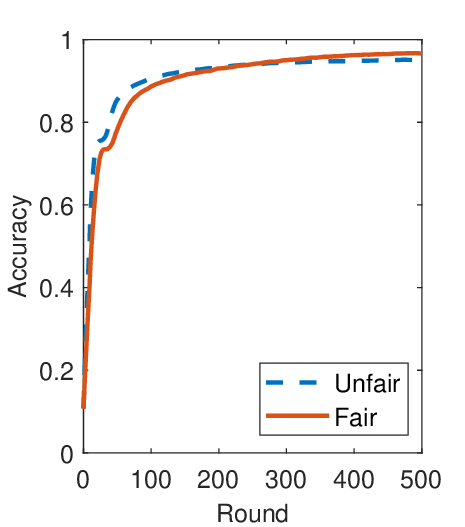}
        \caption{Degree $3$ (MNIST)}
        \label{fig:pre-fairness-impact-mnist-mislabel-80-3}
    \end{subfigure}%
    \begin{subfigure}{0.24\linewidth}
        \centering
        \includegraphics[width=\textwidth]{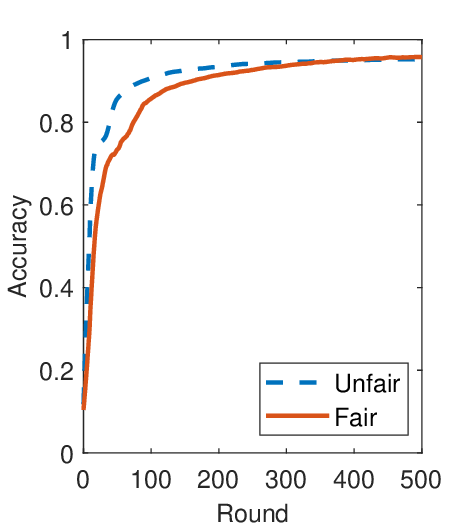}
        \caption{Degree $5$ (MNIST)}
        \label{fig:pre-fairness-impact-mnist-mislabel-80-5}
    \end{subfigure}
    \begin{subfigure}{0.24\linewidth}
        \centering
        \includegraphics[width=\textwidth]{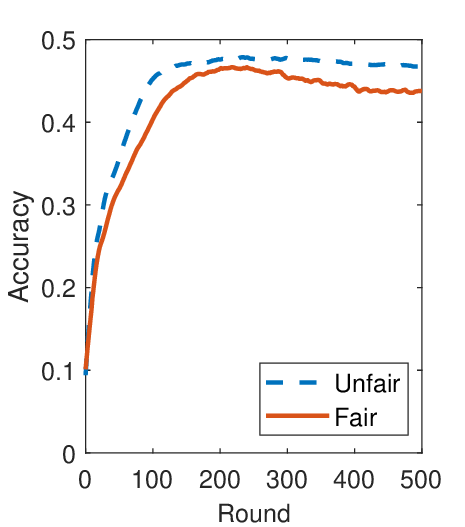}
        \caption{Degree $3$ (CIFAR-10)}
        \label{fig:pre-fairness-impact-cifar10-mislabel-80-3}
    \end{subfigure}%
    \begin{subfigure}{0.24\linewidth}
        \centering
        \includegraphics[width=\textwidth]{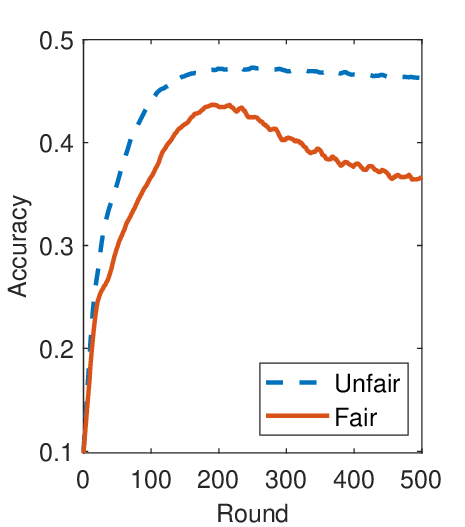}
        \caption{Degree $5$ (CIFAR-10)}
        \label{fig:pre-fairness-impact-cifar10-mislabel-80-5}
    \end{subfigure}   
	\normalsize
    \caption{Test accuracy over training rounds under fair vs. unfair client selection with 80\% sequential mislabeling}
    \label{fig:pre-fairness-mislabel}	
\end{figure}

For mislabeled samples, we applied sequential mislabeling with degrees $3$ and $5$, respectively, to 80\% of the clients.
For this setting, 
the Unfair policy excluded clients with mislabeled samples when randomly selecting clients for participation. The results are shown in Fig.~\ref{fig:pre-fairness-mislabel}.
	
The experiment results shown in Figs.~\ref{fig:pre-fairness-impact-cifar10-mislabel-80-3} and \ref{fig:pre-fairness-impact-cifar10-mislabel-80-5} revealed that random client selections significantly reduced performance compared with the Unfair policy, which is different from the results with the label-skew setting. 
The results suggest that mislabeled samples did not contribute valuable learning information.

\section{Evaluating Datasets for Client Selection}
Given the importance of privacy in FL, the server is strictly prohibited from accessing any distributional information from the local datasets, including label and feature distributions.
Dataset size is the only exception, as it reveals little about the content or distribution of a client's private data.
This restriction calls for a robust evaluation mechanism for client selection that can accurately assess each client's contribution without breaching privacy. 
In the following, we introduce a method to evaluate the quality of the clients' dataset under privacy requirements.


\subsection{Datasize Score}
We define $C$ to be the set of FL clients.
Let $D_i$ be the dataset of client $i$. Client $i$'s Datasize Score is $d_i=1$ if all clients have equal dataset sizes. Otherwise,
\begin{align}
    d_i = \frac{|D_i| - \min_{j \in C}\{|D_j|\}}{\max_{j \in C}\{|D_j|\} - \min_{j \in C}\{|D_j|\}},
    \label{eq:data size score}
\end{align}
which is the normalized size of each client's dataset, ranging from 0 to 1.

\subsection{Quality Score}
We use \emph{Quality Score} to capture the joint impact of mislabeling and label skew. 
Relying on clients to self-report dataset analyses is problematic for three reasons. First, clients may lack a complete or globally consistent view of their own data distributions. Second, clients may be unwilling to share such analyses due to privacy concerns. Third, and most critically, clients may deliberately manipulate their reported analyses to increase their chances of being selected. These concerns together motivate a server-based assessment approach.
As suggested in \cite{DLR+21}, 
we take a server-based assessment where, at the end of each training round, the server evaluates the quality of each client's uploaded model using the server's own test dataset.
The test accuracy serves as an indicator of the client's dataset quality. 
Let $\tilde{\text{acc}_i}$ denote client $i$'s test accuracy.
The Quality Score of client $i$'s dataset is
\begin{align}
    q_i = \frac{\tilde{\text{acc}_i} - \min_{j \in C}\{\tilde{\text{acc}_j}\}}{\max_{j \in C}\{\tilde{\text{acc}_j}\} - \min_{j \in C}\{\tilde{\text{acc}_j}\}}.
    \label{eq:battery_constraint}
\end{align} 
This scaling ensures that the client with the lowest test accuracy scores 0, while the highest scores 1.

We conducted two experiments to validate the effectiveness of the Quality Score.
\subsubsection{Label-Skew Evaluation}
We used 100 clients, each with a dataset of 500 samples.  
The first ten clients had only one label. The number of unique labels increased by one for every subsequent group of 10 clients. The local training epochs for evaluation were set to 30.

\begin{figure}[tb]
    \centering
    \begin{subfigure}{0.32\linewidth}
        \centering
        \includegraphics[width=1\linewidth]{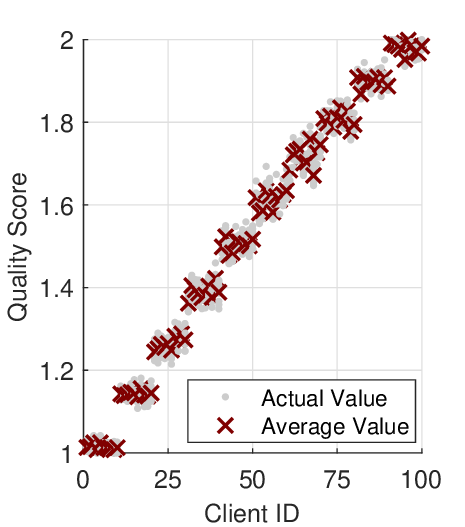}
        \caption{MNIST, 500 samples}
        \label{fig:mnist-skew-impact}			
    \end{subfigure}
    \begin{subfigure}{0.32\linewidth}
        \centering
        \includegraphics[width=\textwidth]{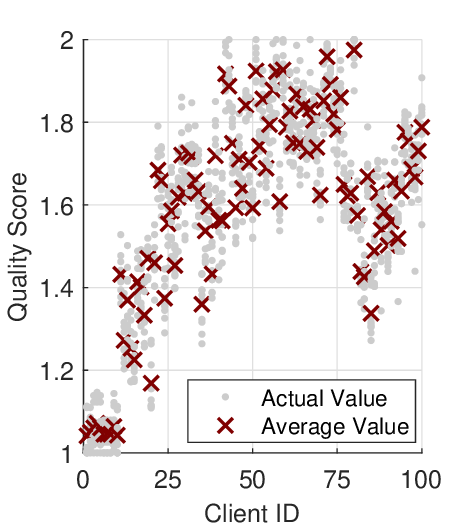}
        \caption{CIFAR-10, 500 samples}
	    \label{fig:cifar10-skew-impact}		
    \end{subfigure}
    \begin{subfigure}{0.32\linewidth}	
        \centering
        \includegraphics[width=\textwidth]{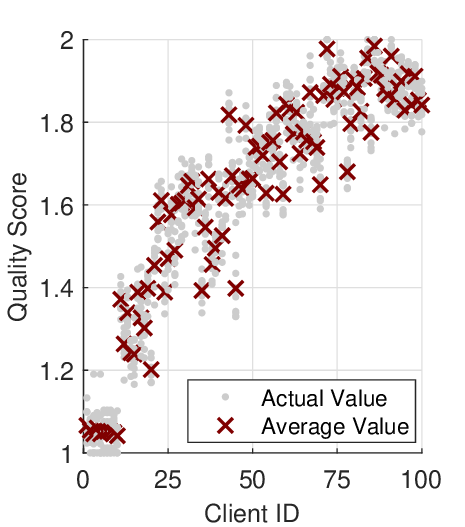}
       \caption{CIFAR-10, 1000 samples}
        \label{fig:cifar10-skew-impact-1000}	   
    \end{subfigure}
    \caption{Quality Score under the label-skew setting}
    \label{fig:pre_eval_quality}	
\end{figure}

The results in Figs.~\ref{fig:mnist-skew-impact} and \ref{fig:cifar10-skew-impact} show that Quality Score increased with the client ID, where the clients with higher IDs possessed more labels. However, there was a performance drop in the CIFAR-10 dataset for clients ranging from IDs 70 to 80. This decline was because each client's total dataset size was fixed at 500 samples. Consequently, the number of samples per label decreased as the number of labels increased, leading to insufficient samples per label for effective learning. To validate this assumption, we conducted a follow-up experiment with the same settings but increased the number of samples per client to 1000 (Fig.~\ref{fig:cifar10-skew-impact-1000}). This adjustment showed an increasing trend in performance. This confirms our hypothesis that more samples per label enhance learning effectiveness and highlights the importance of dataset size.

\subsubsection{Mislabel Level Evaluation} 
This experiment also included 100 clients, each possessing a dataset of 500 samples. To explore different levels of mislabeling, the datasets for the first ten clients contained correctly labeled data (i.e., zero mislabeling rate). The mislabeling rate increased by 10\% for every subsequent group of 10 clients. The local training epochs were also set to 30.
\begin{figure}[tb]
    \centering
    \begin{subfigure}{0.40\linewidth}
        \centering
        \includegraphics[width=\textwidth]{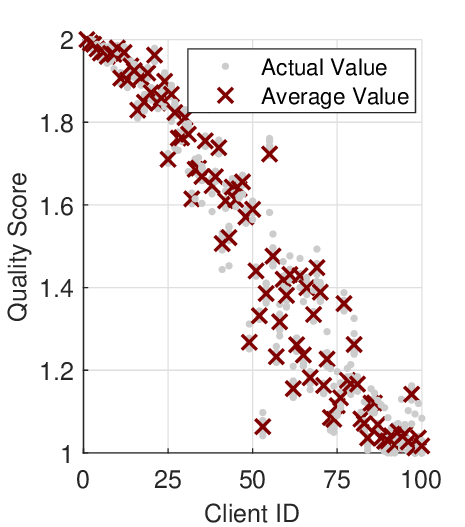}
        \caption{MNIST}
        \label{fig:mnist-mislabel-impact}
    \end{subfigure}%
    \begin{subfigure}{0.40\linewidth}
        \centering
        \includegraphics[width=\textwidth]{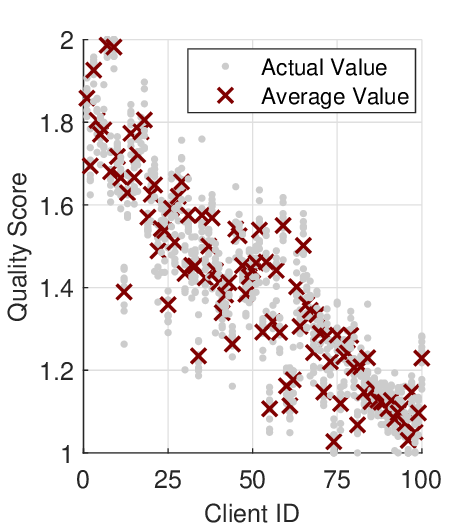}
        \caption{CIFAR-10}
        \label{fig:cifar10-mislabel-impact}
    \end{subfigure}
    \caption{Quality Score under the mislabeling setting}	
	\label{fig:mislabel-impact}		
\end{figure}
 
The results shown in Fig.~\ref{fig:mislabel-impact} exhibit a trend that Quality Score decreased as the client ID increased, whereas the clients with higher IDs suffered from heavier mislabeling.

The results of both experiments confirm the effectiveness of Quality Score in assessing the quality of a client's dataset. 

\subsection{Fairness Score}
We define a client's \emph{Fairness Score} as a non-negative value that reflects its importance when we want to fairly select clients. It is zero for every client initially, increased by $\beta\geq 1$ per round, and decreased by $1$ if the client is selected in the previous round.  
Specifically, let $f^t_i$ be client $i$'s Fairness Score in round $t$, we have
\begin{align}
\label{eq:fti}
        f^t_i =
        \begin{cases}
            0, & t=0,\\
            f^{t-1}_i + \beta - a_i^{t-1}, & \text{otherwise,}
        \end{cases}
\end{align} 
where $\beta$ is a constant and $a_i^{t-1}\in \{0,1\}$ indicates whether client $i$ was selected in round $t-1$.
Consequently, a client will have a high Fairness Score if it has not been selected for a long time. 

Our experiments in Sec.~\ref{pre_exp_fairness} showed that when label skew is the primary source of data heterogeneity, fairness in client selection significantly improves both accuracy and convergence speed. However, when clients have heavily mislabeled datasets, enforcing fairness can harm performance by including low-quality clients.
Ideally, the adaptive mechanism should respond differently to these two sources of heterogeneity. However, since Quality Score captures their joint impact and cannot distinguish between them, we design the mechanism to prioritize quality over fairness when Quality Score variance is high, which is the more conservative choice given that mislabeled samples provide no learning benefit.

The proposed mechanism adjusts the importance of fairness dynamically.
In each round $t$, let $C^t_{\text{ava}}$ denote the subset of $C$ that are available for training.
We select 
the top $L^t_\text{fair}$ clients from $C^t_\text{ava}$ with the highest Fairness Scores. 
The value of $L^t_\text{fair}$ is 
\begin{align}
\label{eq:fair_select}
\scalemath{0.95}{L^t_{\text{fair}} = \max\left(N_\text{max}, |C^t_{\text{ava}}|\bigl[\alpha_1 \frac{\sum_{i \in C^t_\textnormal{ava}} (q_i - \overline{q})^2}{|C^t_{\text{ava}}|}  + \alpha_2\bigr]\right)},
\end{align}
where $N_\text{max}$ is the maximum number of clients to be selected, $q_i$ is client $i$'s Quality Score, and $\overline{q}=\sum_{i \in C^t_\textnormal{ava}} q_i/|C^t_\textnormal{ava}|$ is clients' mean Quality Score in $C^t_\text{ava}$. 
Let $C^t_\text{fair}\subseteq C^t_{\text{ava}}$ be the set of selected clients. 
Parameter $\alpha_2$ ensures a minimum proportion of available clients are included in $C^t_\text{fair}$, while $\alpha_1$ controls how sensitively $L^t_\text{fair}$ expands in response to the variance of Quality Scores among available clients.

When $C^t_{\text{ava}}$ has a small variance of Quality Score, indicating that neither mislabeling nor label skew is significant among available clients, 
the mechanism recruits few clients with high Fairness Scores in $C^t_\text{fair}$, effectively prioritizing fairness. 
When the variance of Quality Score is high, indicating significant data heterogeneity among clients, the mechanism 
 expands the candidate pool so that clients with high Quality Scores can still be selected regardless of their Fairness Scores. Note that this design prioritizes quality over fairness under high heterogeneity, which is the correct behavior when mislabeling is the dominant source of variance but may be suboptimal when label skew dominates, since Sec.~\ref{pre_exp_fairness} showed that fairness benefits model performance in the label-skew setting.



\section{Conclusions}
This study comprehensively evaluated the impacts of imperfect datasets (specifically, quantity skew, label skew, and mislabeled samples) on model accuracy and convergence speed in FL, and experimentally showed that the optimal client selection strategy differs depending on the dominant source of heterogeneity: fairness should be emphasized under label skew but de-emphasized when mislabeling dominates. To assess client contributions without breaching privacy, we proposed a scoring mechanism comprising a Datasize Score, a Quality Score derived from server-based model evaluation, and a Fairness Score, with an adaptive mechanism that dynamically balances fairness and data quality based on Quality Score variance. A limitation of the current design is that Quality Score captures the joint impact of label skew and mislabeling but cannot distinguish between them, since high variance may arise from either source. Developing separate metrics to quantify these two factors independently is an important direction for future work.

\section*{Acknowledgment}
This work was supported in part by the National Science and Technology Council of Taiwan under grant numbers NSTC 114-2218-E-A49-017 and NSTC 114-2218-E-A49-018.

\bibliographystyle{IEEEtran}
\bibliography{ref}

@InProceedings{MMR+17,
  author    = {H. Brendan McMahan and Eider Moore and Daniel Ramage and Seth Hampson and Blaise Agüera y Arcas},
  booktitle = {Proc. 20th Int'l Conf. on Artificial Intelligence and Statistics},
  title     = "Communication-efficient learning of deep networks from decentralized data",
  year      = 2017,
  month     = Apr,
  pages     = "1273-1282",
}

@article{ZLL+18,
  author = {Zhao, Yue and Li, Meng and Lai, Liangzhen and Suda, Naveen and Civin, Damon and Chandra, Vikas},  
  title = "Federated Learning with Non-{IID} Data",
  journal={arXiv:1806.00582},
  year = {2018}
}

@InProceedings{CWJ22,
  title = "Towards Understanding Biased Client Selection in Federated Learning",
  author =    "{Yae Jee} Cho and Jianyu Wang and Gauri Joshi",
  booktitle = "Proc. 25th Int'l Conf. on Artificial Intelligence and Statistics",
  pages = 	 "10351-10375",
  year = 	 2022,
  volume = 	 {151},
  month = 	 mar
}

@article{SLS+23,
      title={Fairness-Aware Client Selection for Federated Learning}, 
      author={Yuxin Shi and Zelei Liu and Zhuan Shi and Han Yu},
      year=2023,
      journal={arXiv:2307.10738}
}

@article{MLL+23,
  title="An optimization method for non-{IID} federated learning based on deep reinforcement learning",
  author={Meng, Xutao and Li, Yong and Lu, Jianchao and Ren, Xianglin},
  journal={Sensors},
  volume={23},
  number={22},
  year={2023}
}

@article{mnist,
    title="{MNIST} handwritten digit database",
    author={LeCun, Yann and Cortes, Corinna and Burges, CJ},
    journal={ATT Labs. Available: http://yann.lecun.com/exdb/mnist},
    year={2010}
}

@techreport{cifar10,
    title={Learning Multiple Layers of Features from Tiny Images},
    author={Krizhevsky, Alex},
    year={2009},
    institution={University of Toronto},
}

@inproceedings{ZXL+21,
  title="Federated Learning on Non-{IID} Data Silos: An Experimental Study",
  author={Qinbin Li and Yiqun Diao and Quan Chen and Bingsheng He},
  booktitle={IEEE 38th Int'l Conf. on Data Engineering},
  year=2022,
  month=may
}

@article{HH+22,
  title={Toward Data Heterogeneity of Federated Learning},
  author={Yuchuan Huang and Chen Hu},
  journal={arXiv:2212.08944},
  year={2022},
  month=dec
}

@inproceedings{DLH+23,
  title={Towards Addressing Label Skews in One-Shot Federated Learning},
  author={Yiqun Diao and Qinbin Li and Bingsheng He},
  booktitle={Proc. Int'l Conf. on Learning Representations},
  year={2023},
}

@article{MHM+20,
  title={Multi-Task Federated Learning for Personalised Deep Neural Networks in Edge Computing},
  author={Jed Mills and Jia Hu and Geyong Min},
  journal={IEEE Trans. Parallel Distrib. Syst.},
  volume={33},
  number={3},
  pages={630--641},
  year={2022},
  month=mar
}

@article{LXS+22,
  title={Multi-Center Federated Learning: Clients Clustering for Better Personalization},
  author={Guodong Long and Ming Xie and Tao Shen and Tianyi Zhou and Xianzhi Wang and Jing Jiang and Chengqi Zhang},
  journal={World Wide Web},
  volume={26},
  pages={481-500},
  year={2023},
  month=feb
}

@article{HCZ+20,
  title={Personalized Cross-Silo Federated Learning on Non-{IID} Data},
  author={Huang, Yutao and Chu, Lingyang and Zhou, Zirui and Wang, Lanjun and Liu, Jiangchuan and Pei, Jian and Zhang, Yong},
  journal={Proc. AAAI Conf. on Artificial Intelligence},
  volume={35},
  number={9},
  pages={7865--7873},
  year={2020},
  month=jul
}

@inproceedings{IJX+24,
  title="Fed{C}lust: Optimizing Federated Learning on Non-{IID} Data through Weight-Driven Client Clustering",
  author={Md Sirajul Islam and Simin Javaherian and Fei Xu and Xu Yuan and Li Chen and Nian-Feng Tzeng},
  booktitle={IEEE Int'l Parallel and Distributed Processing Symp. Workshops},
  year={2024},
  month=may,
  address={San Francisco, CA, USA}
}

@inproceedings{NNN+22,
  title="{FedDRL}: Deep Reinforcement Learning-based Adaptive Aggregation for Non-{IID} Data in Federated Learning",
  author={Nang Hung Nguyen and Phi Le Nguyen and Thuy Dung Nguyen and Trung Thanh Nguyen and Duc Long Nguyen and Thanh Hung Nguyen and Huy Hieu Pham and Thao Nguyen Truong},
  booktitle={Proc. ICPP},
  year={2023}
}

@article{DLR+21,
  title="{AUCTION}: Automated and Quality-Aware Client Selection Framework for Efficient Federated Learning",
  author={Yongheng Deng and Feng Lyu and Ju Ren and Huaqing Wu}, 
  journal={IEEE Trans. on Parallel and Distributed Systems},
  volume={33},
  number={8},
  pages={1996-2009},
  year={2022},
  month=aug,
  publisher={IEEE}
}

@article{ZLT+22,
  title={Deep Reinforcement Learning Based Scheduling Strategy for Federated Learning in Sensor-Cloud Systems},
  author={Tinghao Zhang and Kwok-Yan Lam and Jun Zhao},
  journal={Future Gener. Comput. Syst.},
  volume={144},
  pages={219-229},
  year={2023},
  month=jul,
  publisher={Elsevier}
}

@article{YPB+20,
  title={Robust Federated Learning With Noisy Labels},
  author={Seunghan Yang and Hyoungseob Park and Junyoung Byun and Changick Kim},
  journal={IEEE Intell. Syst.},
  volume={37},
  number={2},
  pages={35--43},
  year={2022},
  month={Mar.-Apr.}
}

@article{LLC+23,
  title={Fed{D}iv: Collaborative Noise Filtering for Federated Learning with Noisy Labels},
  author={Jichang Li and Guanbin Li and Hui Cheng and Zicheng Liao and Yizhou Yu},
  journal={arXiv:2312.12263},
  year={2024},
  month=feb,
  version={v3},
}

@inproceedings{XCQ+22,
  title={Fed{C}orr: Multi-Stage Federated Learning for Label Noise Correction},
  author={Jingyi Xu and Zihan Chen and Tony Q.S. Quek and Kai Fong Ernest Chong},
  booktitle={Proc. CVPR},
  year={2022},
  month=jun
}

@article{ZYC+22,
  title={{CLC}: A Consensus-based Label Correction Approach in Federated Learning},
  author={Bixiao Zeng and Xiaodong Yang and Yiqiang Chen and Hanchao Yu and Yingwei Zhang},
  journal={ACM Trans. Intell. Syst. Technol.},
  volume={13},
  number={5},
  article={75},
  pages={1-23},
  year={2022},
  month=jun
}

@article{YQW+21,
  title={Client Selection for Federated Learning With Label Noise},
  author={Miao Yang and Hua Qian and Ximin Wang and Yong Zhou and Hongbin Zhu},
  journal={IEEE Trans. Veh. Technol.},
  volume={71},
  number={2},
  pages={2193-2197},
  year={2022},
  month=feb
}

@article{FYB+20,
  title={Mitigating Sybils in Federated Learning Poisoning},
  author={Clement Fung and Chris J.M. Yoon and Ivan Beschastnikh},
  journal={arXiv:1808.04866},
  year={2020},
  month=jul,
  version={v5},
}

@article{TSO+24,
  title={Labeling Chaos to Learning Harmony: Federated Learning with Noisy Labels},
  author={Vasileios Tsouvalas and Aaqib Saeed and Tanir Ozcelebi and Nirvana Meratnia},
  journal={ACM Trans. Intell. Syst. Technol.},
  volume={15},
  number={2},
  article={22},
  pages={1-26},
  year={2024},
  publisher={ACM},
  month=feb
}

@article{SYL+21,
  title={Towards Fairness-Aware Federated Learning},
  author={Yuxin Shi and Han Yu and Cyril Leung},
  journal={IEEE Trans. Neural Netw. Learn. Syst.},
  pages={11922--11938},
  volume={35},
  number={9}, 
  year={2024},
  month=sep
}

@article{HLW+21,
  title={An Efficiency-Boosting Client Selection Scheme for Federated Learning With Fairness Guarantee},
  author={Tiansheng Huang and Weiwei Lin and Wentai Wu and Ligang He and Keqin Li},
  journal={IEEE Trans. Parallel Distrib. Syst.},
  volume={32},
  number={7},
  pages={1552--1564},
  year={2021},
  month=jul
}

@article{BLM+22,
  title={Lyapunov-Based Optimization of Edge Resources for Energy-Efficient Adaptive Federated Learning},
  author={Claudio Battiloro and Paolo Di Lorenzo and Mattia Merluzzi and Sergio Barbarossa},
  journal={IEEE Trans. Green Commun. Netw.},
  volume={7},
  number={1},
  pages={265-280},
  year={2023},
  month=mar,
  publisher={IEEE}
}

@article{ZZQ+22,
  title={Online Client Selection for Asynchronous Federated Learning With Fairness Consideration},
  author={Hongbin Zhu and Yong Zhou and Hua Qian and Yuanming Shi and Xu Chen and Yang Yang},
  journal={IEEE Trans. Wireless Commun.},
  volume={22},
  number={4},
  pages={2493-2506},
  year={2023},
  month=apr,
  publisher={IEEE}
}

@article{HLS+21,
  title={Stochastic Client Selection for Federated Learning With Volatile Clients},
  author={Tiansheng Huang and Weiwei Lin and Li Shen and Keqin Li and Albert Y. Zomaya},
  journal={IEEE Internet Things J.},
  volume={9},
  number={20},
  pages={20055--20070},
  year={2022},
  month=oct,
  publisher={IEEE}
}

\end{document}